\pdfoutput=1

\documentclass[11pt]{article}

\usepackage{ACL2023}

\usepackage{times}
\usepackage{latexsym}

\usepackage[T1]{fontenc}

\usepackage[utf8]{inputenc}

\usepackage{microtype}

\usepackage{inconsolata}

\usepackage{hyperref}
\usepackage{url}

\usepackage{times}
\usepackage{latexsym}
\usepackage{subcaption}
\usepackage{comment}
\usepackage{times}
\usepackage{latexsym}
\usepackage{amsmath}
\usepackage{multirow}
\usepackage{multicol}
\usepackage{graphicx}
\usepackage{CJKutf8}
\usepackage{float}
\usepackage{arydshln}
\usepackage{amsfonts}
\usepackage{multicol}
\usepackage{comment}
\usepackage[font=small,labelfont=bf]{caption}
\title{On the Cross-lingual Effect of Sharing with Subnetworks}
\title{on the modularity of multilingual lms and the effect of sparse fine-tuning}
\title{On the Cross-lingual Effect of Sparse Fine-tuning with Subnetworks: Shortcomings and Promising Directions}
\title{On the Cross-lingual Effect of Sparse Fine-tuning with Subnetworks}
\title{Understanding Cross-lingual Sharing\\ with Language-Specific Subnetworks}
\title{Investigating Cross-lingual Sharing\\ with Language-Specific Subnetworks}

\title{Examining Language-Specific Subnetworks\\to Understand Cross-lingual Sharing}
\title{Understanding Cross-lingual Sharing \\ through Subnetworks}
\title{Examining Modularity in Multilingual LMs \\ through  their Subnetworks}
\title{Examining Subnetworks\\to Understand Cross-lingual Sharing}

\title{Examining Subnetworks to Understand\\ Modularity and Cross-lingual Sharing}

\title{Examining Subnetworks for Studying\\ Modularity and Cross-lingual Sharing}
\title{Examining Language Specialization of Subnetworks in Multilingual LMs}
\title{Examining Modularity in Multilingual LMs\\ via Language-Specialized Subnetworks}

\author{Rochelle Choenni
  \\
  University of Amsterdam
  \\
  \texttt{r.m.v.k.choenni@uva.nl}
   \And
  Ekaterina Shutova
  \\
  University of Amsterdam
  \\
  \texttt{e.shutova@uva.nl}
   \And
  Dan Garrette 
  \\
  Google DeepMind
  \\
  \texttt{dhgarrette@google.com}
}

\begin{document}
\maketitle

\begin{abstract}
Recent work has proposed explicitly inducing language-wise modularity in multilingual LMs via sparse fine-tuning (SFT) on per-language subnetworks as a means of better guiding cross-lingual sharing.
In this work, we investigate 
(1)~the degree to which language-wise modularity \emph{naturally} arises within models with no special modularity interventions, and 
(2)~how cross-lingual sharing and interference differ between such models and those with explicit SFT-guided subnetwork modularity. 
To quantify language specialization and cross-lingual interaction, we use a Training Data Attribution method that estimates the degree to which a model's predictions are influenced by in-language or cross-language training examples.
Our results show that language-specialized subnetworks do naturally arise, and that SFT, rather than always increasing modularity, can decrease language specialization of subnetworks in favor of more cross-lingual sharing.
\end{abstract}

\section{Introduction}
Multilingual language models (LMs) can achieve remarkable performance across many languages thanks to phenomena like cross-lingual sharing \citep{pires2019multilingual}, but they still suffer from the ``curse of multilinguality'' \citep{conneau2020unsupervised} as performance can be hindered by negative cross-language interference \citep{wang2020negative}. 
Recently, new methods have been proposed for mitigating these negative effects by training specialized model components for processing individual languages \citep{pfeiffer2022lifting}. These approaches, which add explicit \textbf{modularity} to the model, are also effective in promoting positive transfer and increasing interpretability \citep{pfeiffer2023modular}.

While previous work has focused on developing techniques for explicitly adding modularity to models, we take a step back and ask: To what degree does language-wise modularity \emph{naturally} arise within a model with no targeted modularity interventions?
To investigate this question, we make use of a method inspired by the Lottery Ticket Hypothesis \citep{frankle2018lottery, chen2020lottery}: 
for each language, we identify a \textbf{subnetwork}---a subset of model parameters---such that when fine-tuned on in-language data, it performs on par with the full model on that language \citep{wang2020negative, nooralahzadeh2020zero}. 
We then use these subnetworks to quantify language-wise modularity in a model by measuring the degree to which the subnetworks depend solely on in-language training examples when making predictions, which we refer to as \emph{language specialization}.
Subnetworks are an appealing method for our study because they do not require the introduction of additional model parameters, which means that we are able to use this approach on a model that has not been explicitly modified to add modularity.

\begin{figure}[!t]
    \centering
\includegraphics[width=0.98\linewidth]{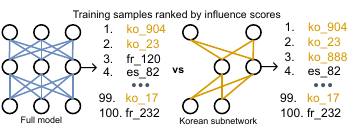}
    \caption{
    We study how in-language training data reliance changes for individual test languages when using a subnetwork compared to the full model at test time. For instance, will a Korean subnetwork rely more on Korean training samples when making a prediction for a Korean test sample?}
    \label{fig:rank_ex}
\end{figure}

Moreover, subnetworks have also proven to be a popular modularization technique because when used to restrict parameter updates as a form of sparse fine-tuning (\textbf{SFT}), they are able to guide cross-lingual sharing toward positive transfer and away from negative interference~\citep{lin2021learning, lu2022language, xu2022s4, choenni2022data, hendy2022domain}. However, less is known about precisely what effects SFT has on the underlying model behavior. Thus, we investigate the following set of questions: (1) To what degree does language-wise modularity naturally arise within a model, when it is not explicitly enforced by restricting gradient updates? (2) How do cross-lingual sharing and interference differ between models without modularity interventions versus SFT-guided language-wise modularity?
(3) Does SFT result in increased language specialization? 
(4) How does the degree of language specialization affect model performance?, and (5) To what degree does the similarity of language-specific subnetworks dictate cross-language influence?

To quantify cross-language interaction, we follow \citet{choenni2023languages} in using a Training Data Attribution (TDA) method, TracIn \citep{pruthi2020estimating}, which measures the degree of influence each training example has on a particular model prediction. By examining the influence each language's training set has on the test predictions for individual languages, we can estimate how much influence languages on average exert cross-lingually.

We conduct experiments on 
three text classification tasks---natural language inference, paraphrasing, and sentiment analysis. For each task, even without special modularity interventions, we are able to identify subnetworks 
that rely more heavily on in-language data than the full model does. 
Additionally, we find that SFT does not always increase this modularity, but instead can decrease language specialization within the subnetworks and boost cross-lingual sharing to improve performance. 
Finally, we provide additional analysis on factors that affect cross-language influence, and find interesting correlations between subnetwork similarity and the amount of positive influence across languages. 


\section{Background and Related work}

\subsection{Modular deep learning}
Modular approaches existed before the rise of pre-trained LMs \citep{shazeer2016outrageously, andreas2016learning}, but have recently regained popularity in NLP. The idea is that modular systems will allow us to improve performance in an interpretable way as modularity provides a more intuitive path to compositionality. 
Various methods have been proposed to implement specialized modules, for instance, by inserting adapter layers into the model \citep{rebuffi2017learning, rebuffi2018efficient, houlsby2019parameter, pfeiffer2022lifting}, replacing fine-tuning by prefix-tuning \citep{li2021prefix}, or by SFT with subnetworks \citep{sun2020learning}. While the former two aim to create modularity \emph{post-hoc} by injecting task-specific parameters into the existing model, the latter approach aims to induce it into the model as an inductive bias during fine-tuning. In this work, we delve deeper into the effects of SFT to understand whether it is able to produce more modular systems. 
While some work studies modularity in both vision and language models \citep{csordas2020neural, zhang2023emergent, lepori2023break, dobs2022brain}, we are the first to study it in multilingual LMs, and to study subnetwork interaction by directly looking at the training data.


\subsection{Subnetworks and SFT}
\citet{frankle2018lottery} showed that subnetworks can be found through pruning methods \citep{han2015learning, li2016pruning} that match the performance of the full model. 
Since then, it has been shown that such subnetworks exist within BERT models \citep{prasanna2020bert, budhraja2021prunability, li2022probing}
 , and that both language-neutral and language-specific subnetworks can be found in multilingual LMs \citep{foroutan2022discovering}.
Hence, sparse training gained popularity in multilingual NLP:  \citet{nooralahzadeh2023improving} show that training \textit{task-specific} subnetworks can help in cross-lingual transfer, \citet{lin2021learning} use \textit{language-pair-specific} subnetworks for neural machine translation, and \citet{hendy2022domain} use \textit{domain-specific} subnetworks. Finally, \citet{ wang2020negative, lu2022language, choenni2022data, xu2022s4} use language-specific subnetworks to improve cross-lingual performance on a range of tasks, e.g.\ speech recognition, dependency parsing and natural language understanding, suggesting that sparse training can reduce negative interference and/or stimulate positive knowledge transfer. While \citet{choenni2022data} found evidence of the former through 
fewer gradient conflicts during training \citep{yu2020gradient}, 
we are the first to study the effect of SFT on cross-lingual data sharing. 

\subsection{Training Data Attribution}
 TDA methods 
 aim to identify a set of training examples that most informed a particular test prediction. Typically, the influence of training point $z_\textit{train}$ on test point $z_\textit{test}$ is formalized as the change in the loss that would be observed for $z_\textit{test}$ if sample $z_\textit{train}$ was omitted during training \citep{koh2017understanding}. Thus, we can use it as a measure of how important $z_\textit{train}$ is for making a prediction for $z_\textit{test}$. TDA methods have been used in NLP for unveiling data artifacts  \citep{han2022orca}, e.g.,\ to detect outlier data \citep{han2020explaining}, enable instance-specific data filtering \citep{lam2022analyzing}, or to correct erroneous model predictions \citep{meng2020pair, guo2021fastif}. 
Following \citet{choenni2023languages}, we instead employ 
it to study cross-lingual data sharing in LMs. To understand how much influence languages exert cross-lingually, \citet{choenni2023languages} 
quantify cross-language influence during multilingual fine-tuning by the percentage that each language's training data contributes to the most influential training samples for each test language. 
While they study the effects of full model fine-tuning, we employ their framework to study modularity in LMs by testing the data reliance behavior of language-specific subnetworks and the effect that SFT has on this.

\section{Methods}

\subsection{Identifying Subnetworks}\label{ref:subnetworks}

Subnetworks are represented by masks that can be applied to the model to ensure that only a subset of the model's parameters are activated (or updated during training). We follow \citet{prasanna2020bert} in using \emph{structured} masks, treating entire attention heads as units which are always fully enabled or disabled. Thus, for a language $\ell$, its subnetwork is implemented as a binary mask $\xi_\ell \in \{0, 1\}^{H \times L}$, where $H$ and $L$ correspond to the number of attention heads and layers. We aim to find masks for languages that prune away as many heads as possible without harming model performance on a given language (i.e.,\ by pruning away heads that are only used by other languages, or that are unrelated to the task). For this, we apply the procedure introduced by \citet{michel2019sixteen}. Starting from a model that is fine-tuned for a task in language $\ell$, we iterate by repeatedly removing the 10\% of heads with the lowest importance scores $\text{HI}^{(i,j)}_\ell$ ($i$=head, $j$=layer), which is estimated based on the expected sensitivity of the model to the mask variable $\xi^{(i,j)}_\ell$:

\begin{equation}
     \textnormal{HI}^{(i,j)}_\ell = \mathop{\mathbb{E}_{x_\ell \sim X_\ell}} \left| \frac{\partial \mathcal{L}(x_\ell)}{\partial \xi^{(i,j)}_\ell} \right|
\end{equation}
where $X_\ell$ is $\ell$'s data distribution,
$x_\ell$ is a sample from that distribution, and 
$\mathcal{L}(x_{\ell})$ is the loss with respect to the sample. Pruning stops when we 
reach $95$\% of the original model performance. 

\subsection{TracIn: Tracing Influence}\label{sec:influence}

\citet{pruthi2020estimating} propose TracIn, a simple TDA method to approximate influence of a training sample over training. They do this by computing the influence of a training sample $z_i$ on the prediction for a test sample $z_\textit{test}$ as follows:
\begin{equation}
    \mathcal{I}(z_i, z_\textit{test}) = \sum^{E}_{e=1} \nabla_\theta \mathcal{L}(z_i, \theta_e) \cdot  \nabla_\theta \mathcal{L}(z_\textit{test}, \theta_e)
\end{equation}
where $\theta_e$ is the checkpoint of the model at each
training epoch. The intuition behind this method is
to approximate the total reduction in the test
loss $ \mathcal{L}(z_\textit{test}, \theta)$ during the training process when
the training sample $z_i$ is used. This gradient product method reduces the problem to
the dot product between the gradient of the training
loss and the gradient of the test loss. As dominating gradients are a known problem in multilingual NLP \citep{wang2020negative}, we also adopt the simple normalization trick from \citet{barshan2020relatif}, i.e.,\ substituting the dot product operation with cosine similarity, thus normalizing by the norm of the training gradients. Lastly, following \citet{pruthi2020estimating}, we reduce computational costs by pre-computing low-memory sketches of the loss gradients of the training points using random projections, and reuse them to compute randomized unbiased estimators of the influence on different test points \citep{woodruff2014sketching}. See Appendix \ref{app:efficiency} for more details.

\section{Experimental setup}
\subsection{Tasks and datasets}

\paragraph{Natural language inference}
The Cross-Lingual Natural Language Inference (XNLI) dataset \citep{conneau2018xnli} contains premise-hypothesis pairs labeled with 
their relationship: `entailment', `neutral' or `contradiction'. The dataset contains parallel data 
of which the original pairs come from English and were translated to other languages. We use English, French, German, Russian and Spanish portions of the dataset. 

\paragraph{Paraphrasing}
Cross-Lingual Paraphrase Adversaries from Word
Scrambling (PAWS-X) \citep{yang2019paws} requires the model to decide if two sentences are paraphrases of one another. 
PAWS-X contains translated data from PAWS \citep{zhang2019paws}. Part of the development and test sets was translated from English by professionals and the training data was translated automatically. 
We experiment with English, French, German, Korean and Spanish.

\paragraph{Sentiment analysis}
The Multilingual Amazon Review Corpus (MARC) \citep{keung2020multilingual} contains Amazon reviews written by users in various languages. Each record in the dataset contains the review text and title, and a star rating. 
The corpus is balanced across 5 star rating, so that each star rating constitutes 20\% of the reviews in each language. Note that this is a non-parallel dataset. 
We experiment with Chinese, English, French, German and Spanish.

\subsection{Training techniques}

\paragraph{Full model fine-tuning} 
\label{sec:fine-tune}
We fine-tune the full XLM-R model \citep{conneau2020unsupervised} on the concatenation of 2K samples from 5 languages, i.e. 10K samples for each task. For XNLI, we follow \citet{han2020explaining} by performing binary classification “entailment or not” ; for MARC, we collapse 1 and 2 stars into a negative and 4 and 5 stars into a positive review category. Training converges at epoch 4 for XNLI, and at epoch 5 for PAWS-X and MARC, obtaining 78\%, 83\%, and 90\% accuracy on their development sets respectively, for more details see Appendix~\ref{app:fine-tuning}.

\paragraph{Sparse fine-tuning (SFT)}\label{sec:sft}
We sample language-specific batches in random order, and each time restrict parameter updates to only those parameters that are enabled within the respective language's identified subnetwork. We use the subnetworks during fine-tuning by restricting the model both in the forward and backward pass.\footnote{We implement this during backpropagation by multiplying the gradients by the binary subnetwork mask, and passing the masked gradients to the optimizer.  In the forward pass, we simply disable the attention heads.} We ensure that we sample each language equally often. All other fine-tuning details remain the same. 

\subsection{Evaluation}

\paragraph{Computing influence scores} We use 500 random test samples from each language and compute influence scores between each test sample and all 10K training instances. 
For each test sample, we retrieve the top $m$=100 training instances with the largest \emph{positive} and the largest \emph{negative} influence scores and refer to them as the set of most positively and negatively influential samples respectively ($m$=100 was previously found to be optimal \citep{choenni2023languages}). Note that negative cosine similarity between gradients have been referred to as gradient conflicts \citep{yu2020gradient}, and were shown to be indicative
of negative interference in the multilingual setting \citep{wang2020negative}. Moreover, we ensure that the model was able to predict the correct label for all test instances that we compute influence scores for such that we only study the training samples 
that influenced the model to make a correct prediction. Also, as we train on parallel data for XNLI and PAWS-X, the content in our training data is identical across languages, giving each language an equal opportunity to be retrieved amongst the most influential samples. 

\paragraph{Quantifying cross-language influence} After obtaining an influence score ranking over our training set for each test sample, we compute how much each training language contributed to the prediction for the test samples in other languages. We then compare the resulting rankings produced using the full model and an identified subnetwork, see Figure~\ref{fig:rank_ex}. As there can be small differences in performance between the subnetworks and the full model, throughout all experiments, we compare cross-language influence for test samples that both models were able to correctly classify. 

\section{Naturally arising modularity}
\label{sec:modularity}
In this section, we study whether modularity has naturally arisen within a model after multilingual full model fine-tuning. As such, the subnetworks are only applied at test time.

\subsection{How specialized are subnetworks?}
To study the degree to which modularity has naturally arisen after full model fine-tuning, we look for subnetworks that naturally specialize in their respective languages.
We quantify language specialization as the extent to which the subnetworks rely solely on in-language training data when making test-time predictions. Thus, for each test language, we use the pruning procedure explained in Section~\ref{ref:subnetworks} to identify a subnetwork within the fine-tuned model. We then compute influence scores on the fine-tuned model, applying the subnetwork mask corresponding to the language of the test example. 
Finally, we compare the model's reliance on in-language data when using these subnetworks against its reliance when no subnetwork mask is applied (i.e. when predicting with the full model).

\begin{figure}[!t]
     \centering
         \includegraphics[width=\linewidth]{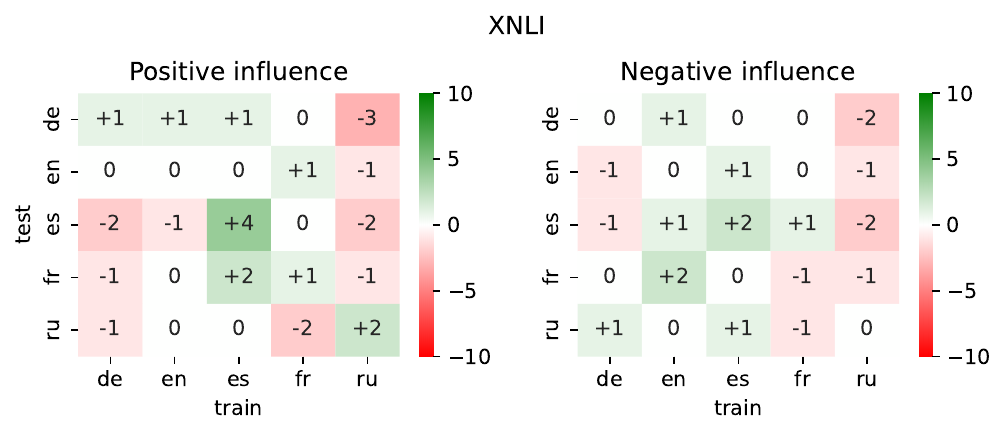}
     \includegraphics[width=\linewidth]{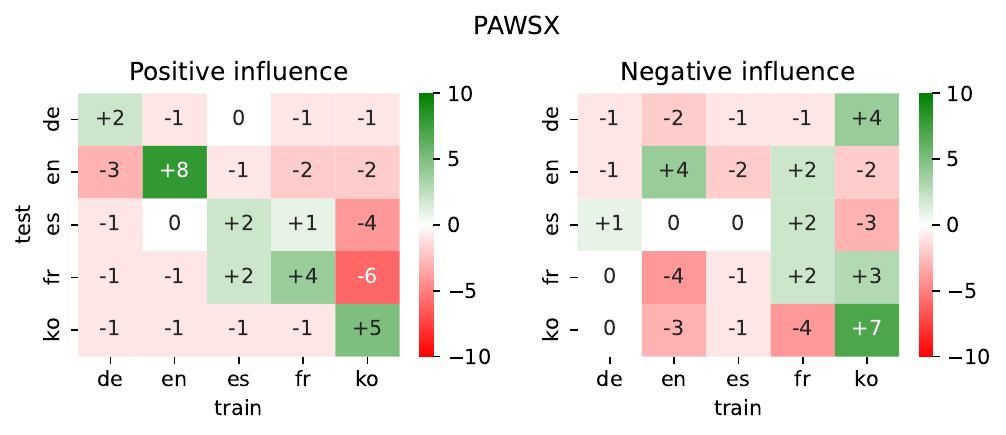}
       \includegraphics[width=\linewidth]{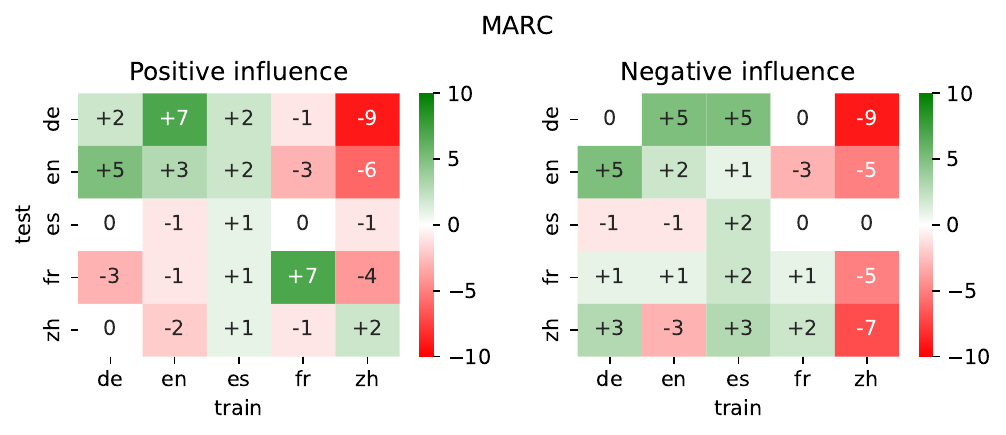}
     \caption{\textbf{(After full model fine-tuning)} The effect of using the identified language-specific subnetwork for each test language compared to the full model at test time. On the $x$-axis we have the training language and on the $y$-axis the test language. The values denote the change (\%) in influence from the training on the test language. Results are averaged over all 500 test samples per language.  }
         \label{fig:diff_subnets}
\end{figure}

\paragraph{Results}
In Figure~\ref{fig:diff_subnets} we show, per task and test language, the change in contribution (\%) to the top 100 most positively and negatively influential samples when using the subnetworks compared to the full model. On the diagonals, we clearly see that for all languages across all tasks, using the subnetwork does mostly result in more positive influence from the respective language (from +1 to +8\%). This indicates that we are able to identify language-specialized subnetworks that are more biased toward relying on in-language data, and thus suggests that some form of modularity naturally exists within the model. For baseline results from the full model and more details on the subnetworks, see Appendices \ref{sec:full_baseline}\footnote{Our results using 500 test samples per language are similar to those from the original paper by \citet{choenni2023languages}.} and \ref{app:subnet_details} respectively.

The effects are less clear when looking at negative influence; here we see that using a language's subnetwork can also decrease negative influence coming from in-language data (e.g.\ Chinese for MARC).
Finally, results from XNLI are overall weaker than for the other tasks. This is in line with results from the full model that showed that, for XNLI, the model relies to the least extent on in-language data, hence we can expect language-specificity to be less strong for these subnetworks. Moreover, for English, we find no difference in language specialization. This can be explained by the fact that the German and Russian subnetworks share 100$\%$ of their capacity with English, making its subnetwork less distinct (see Appendix~\ref{app:subnet_details}).

\begin{figure}[!t]
    \centering
     \includegraphics[width=\linewidth]{ 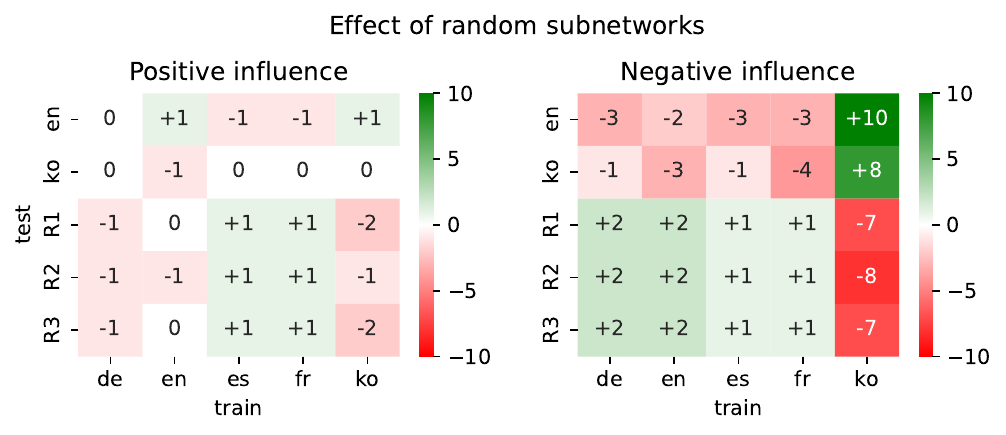}
     \caption{\textbf{(After full model fine-tuning)} The effect on cross-language influence when using random (R) and suboptimal (English and Korean) subnetworks on German as a test language for PAWS-X. }
         \label{fig:random}
\end{figure}

\paragraph{Cross-language influence} We have shown that language-specialized subnetworks rise. We now analyze how cross-language influence differs within such subnetworks compared to the full model. For MARC, we see that the increase in positive self-influence (diagonal) can be smaller than the increase in positive influence from related languages. In particular, we see that using a German subnetwork strongly increases positive influence from the most typologically similar training language, i.e.,\ English (+7$\%$), and vice versa (+5$\%$). While the change in positive influence from related languages is stronger than that of the respective subnetwork's language, the subnetwork still relies more on in-language data when looking at absolute numbers. For German, the full model was relying for 33$\%$ on in-language data, which using its subnetwork increased to 35\% (+2$\%$). Yet, English initially only contributed 17\% to German, which after using its subnetwork increased to 24\% (+7\%)  (see Appendix \ref{sec:full_baseline} for the full model results). 
We suspect that we observe the effect of positive knowledge transfer through cross-lingual sharing here. Similar to the full model, when subnetworks have exploited most useful in-language data, they start benefitting more from exploiting other languages' data instead. 

\subsection{Random and suboptimal subnetworks}
As baselines to our identified subnetworks, we study whether evidence for language specialization can also be found within random and suboptimal subnetworks for PAWS-X. \emph{Random}: we shuffle the binary subnetwork masks with 3 random seeds, and recompute scores from them. Note that we do this only for German---we saw the weakest increase in language-specificity for German (+2\%, see Figure~\ref{fig:diff_subnets}), thus it should be the easiest to get similar results from a random subnetwork. \emph{Suboptimal}: we pick the subnetwork from the most similar and distant language to German, i.e.,\ English and Korean, and recompute influence scores for German (i.e.,\ testing the effect of applying the subnetwork from a language $A$ to a language's $B$'s input.). 

\begin{figure}[!t]
     \centering
         \includegraphics[width=1.01\linewidth]{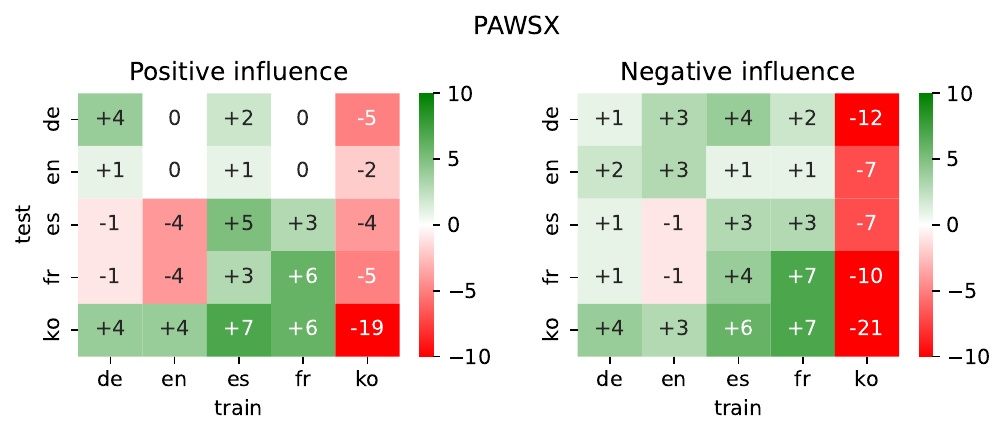}
     \includegraphics[width=1.01\linewidth]{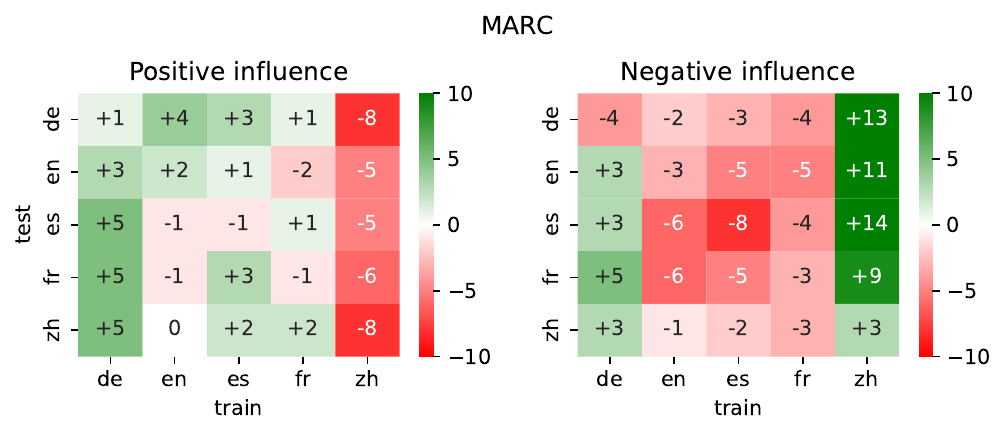}
     \caption{\textbf{(After SFT)} The effect of using the identified language-specific subnetwork for each test language compared to using the full model at test time. On the $x$-axis we have the training language and on the $y$-axis the test language. The values denote the change (\%) in influence from the training on the test language. Results are averaged over all 500 test samples per language  }
         \label{fig:sft_diffs}
\end{figure}

\paragraph{Results} 
In Figure~\ref{fig:random}, we find that using random subnetworks overall causes little change to the score distributions as compared to the full model. In particular, we find that in none of the cases the influence of German increases.  Also, it is evident that the behavior from the suboptimal subnetworks is different from the random subnetworks. For instance, we find that using either the correct English or Korean subnetworks result in a strong increase of negative interference from Korean (+10 and 8\%). Yet, when we use the random subnetworks we instead observe a strong tendency for Korean to decrease in negative influence. These results show that our identified subnetworks encode meaningful differences compared to randomly selected ones. 

  \begin{figure}[!t]
    \centering
      \includegraphics[width=0.9\linewidth]{ 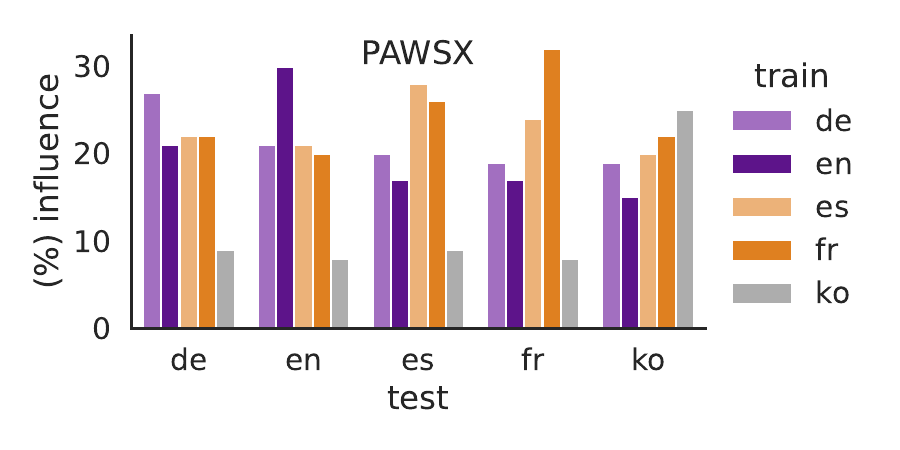}
     \includegraphics[width=0.9\linewidth]{ 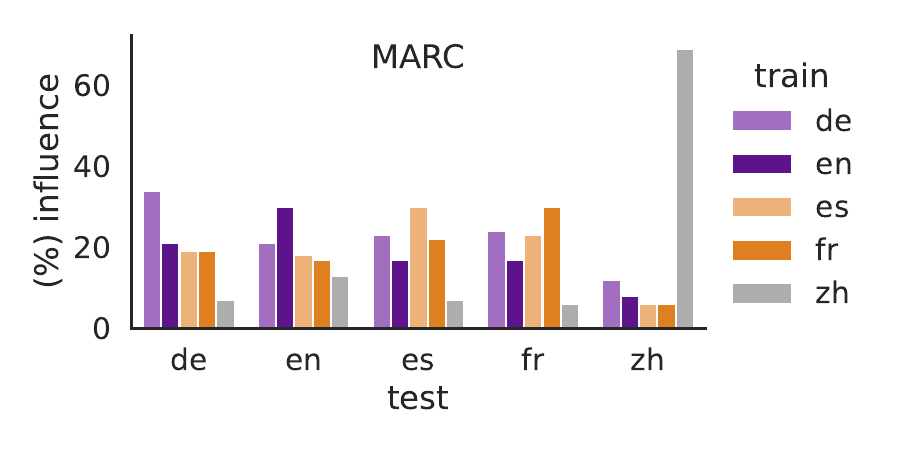}
     \vspace{-0.2cm}
       \caption{The positive influence ($\%$) from each training language on each test language in absolute numbers. The values are retrieved from the subnetworks after SFT. Note that the $y$-axes are not on the same scale.}
         \label{fig:dists}
  \end{figure}

\section{How does SFT affect modularity?}
\label{sec:sft-subs}

In Section~\ref{sec:modularity}, we studied whether modularity had naturally arisen in the model in the form of language-specialized subnetworks. We now study the effect that SFT has on these subnetworks, i.e.,\ does it further encourage modularity within the model? Thus, instead of only applying the subnetworks at test time, as was done in the previous section, we now use the same identified subnetworks, but apply them both during SFT and at test time.
We then recompute influence scores between test and training samples, 
and observe the change in language specialization  compared to full model fine-tuning. This way, we test whether SFT, compared to full model fine-tuning, causes the subnetworks to further specialize on in-language data. 

Given that the subnetworks found for XNLI had the smallest effect on cross-language data reliance, and we did not find a distinct English subnetwork, we conduct further experiments on PAWS-X and MARC (that contain parallel and non-parallel data respectively) to reduce computational costs. 
Also, we confirm that SFT improves performance on both tasks (see Appendix~\ref{app:additional}). 
 For PAWS-X, we obtain an average test accuracy of 74.8$\%$ 
 when using subnetworks after full model fine-tuning and 78.4$\%$ after SFT (+$3.6\%$). For MARC we see an average improvement of +$1.2\%$ when using SFT.

\paragraph{Results}
In Figure~\ref{fig:sft_diffs}, we see the change in language influence compared to using the full model. We find that the in-language data reliance of some subnetworks tends to decrease after SFT (i.e.,\ Korean for PAWS-X and Chinese, French, and Spanish for MARC). This is surprising given that SFT is generally seen as a modularization technique. Whilst it is important to note that all subnetworks still mostly rely on in-language data as shown by the absolute numbers reported in Figure~\ref{fig:dists}, our results suggest that the benefit of SFT cannot fully be attributed to language specialization of the subnetworks. Instead, cross-lingual sharing, guided through subnetwork interaction, is likely a contributing factor as well. Finally, as our results suggest that SFT does not necessarily strengthen language specialization, it sheds doubt on SFT as a method for creating more modular systems.
\begin{figure}[!t]
    \centering
    \includegraphics[width=1\linewidth]{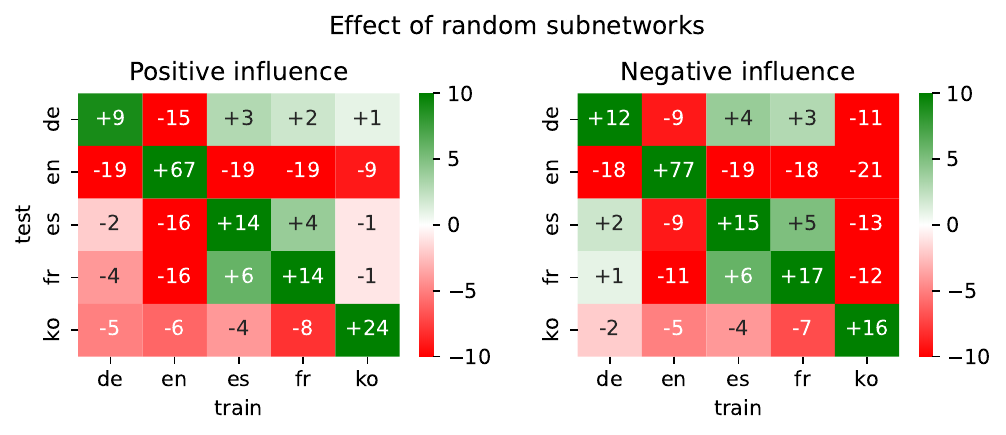}
    \caption{\textbf{(After SFT)} The effect that SFT with \emph{random} subnetworks has for PAWS-X on the amount of language specialization that the subnetworks acquire compared to full model fine-tuning. }
    \label{fig:random-sft}
\end{figure}

\begin{figure}[!t]
    \centering
    \includegraphics[width=0.49\linewidth]{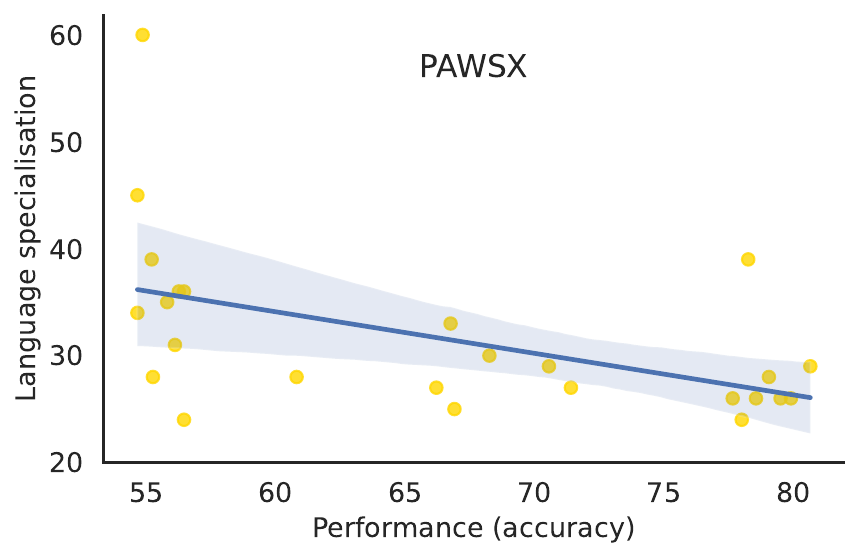}
     \includegraphics[width=0.49\linewidth]{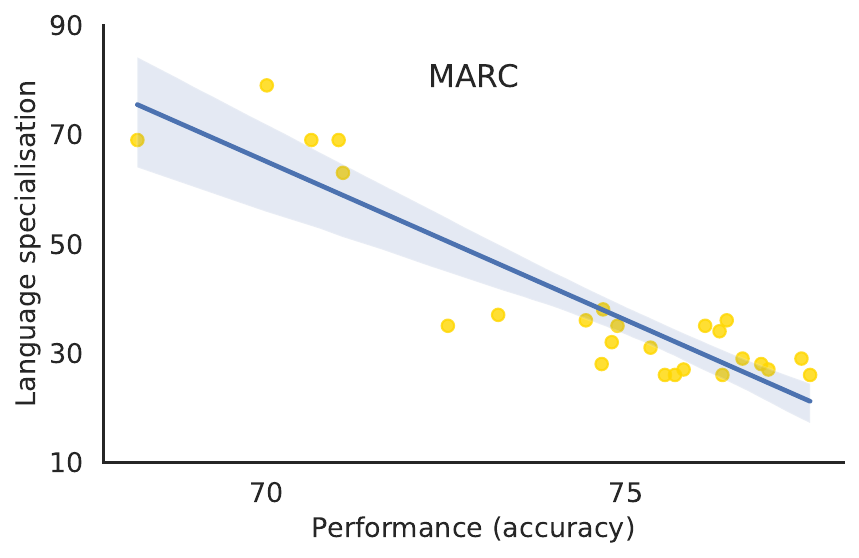}
    \caption{The correlation between language specialization and performance accuracy for PAWS-X and MARC. We compute scores for all languages and model checkpoints.}
    \label{fig:pawsx_corr}
\end{figure}

\subsection{SFT with random subnetworks}\label{sec:rdm_sft}
As a baseline to our previous findings, we now test whether any randomly found subnetwork could in principle be taught to specialize in a language when we use SFT as a training method. Thus, for each language, we shuffle the language-specific subnetworks to obtain a random subnetworks with the same sparsity level. We then use these random subnetworks, both during SFT and at test time, and repeat the procedure from Section~\ref{sec:sft-subs}.

\paragraph{Results}
Surprisingly, in Figure~\ref{fig:random-sft}, we see that random subnetworks to a much larger extent rely on in-language data than the identified subnetworks used in Section~\ref{sec:sft-subs}. In particular, we see that the model barely relies on cross-lingual sharing for English (+64$\%$ compared to the full model, which results in $97\%$ reliance on English data when using the subnetwork). Yet, we also find that these highly specialized subnetworks perform considerably worse, on average only obtaining $\pm$56$\%$ across languages. Given that random subnetworks do not contain the necessary information to process the language, we hypothesize that (1)  during SFT they need to learn both the task and language, which causes them to focus on in-language data first, and (2) cross-lingual sharing will only happen once the in-language data has been fully exploited. 
Our results show that any subnetwork can in principle learn to specialize in one language, but that this might be suboptimal.

\section{Further analysis}
In Section~\ref{sec:sft-subs}, we show that SFT only sometimes causes our identified subnetworks to rely more on in-language data, yet unlike random subnetworks, do seem to encode meaningful information. To understand where the performance improvements from SFT come from, we perform further analysis on how: (1) language specialization correlates with performance, (2) subnetwork similarity affects cross-language influence, and (3) sharing dynamics evolve during full model fine-tuning versus SFT. 
\subsection{Correlation between language specialization and performance}\label{sec:corr} 
We find that SFT only decreases performance on French for PAWS-X (Table~\ref{tab:performance}, Appendix~\ref{app:additional}), which happens to also be the subnetwork that showed the strongest increase in language specialization after SFT (+6$\%$) in Section~\ref{sec:sft-subs}. 
To test to what degree subnetwork performance benefits from language specialization, we study the correlation between the two 
using data from all model checkpoints.

\paragraph{Results}
In Figure~\ref{fig:pawsx_corr}, we see that, for both tasks, stronger language specialization is negatively correlated with model performance. 
This finding further supports our hypothesis that the strength of SFT really comes from cross-lingual sharing that happens between the subnetworks rather than from the language specialization of the subnetworks themselves. Intuitively, this makes sense as SFT forces the model to squeeze information into the smaller subsets of model parameters, which has to improve performance on a set of training languages, and as such, requires better cross-lingual sharing.

\subsection{Correlation between subnetwork similarity and cross-language influence}\label{sec:corr_overlap} 
SFT allows for cross-lingual interaction through subnetwork overlap in which the model parameters are shared between languages. This sharing mechanism is motivated by the idea that similar languages are encoded by similar subnetworks (and thus naturally dictating cross-lingual sharing by their overlap). To test this hypothesis we study the correlation between subnetwork similarity and cross-language influence between language pairs. 
We measure similarity by the cosine similarity between the flattened binary subnetwork masks.

\begin{figure}[!t]
    \centering
    \includegraphics[width=0.49\linewidth]{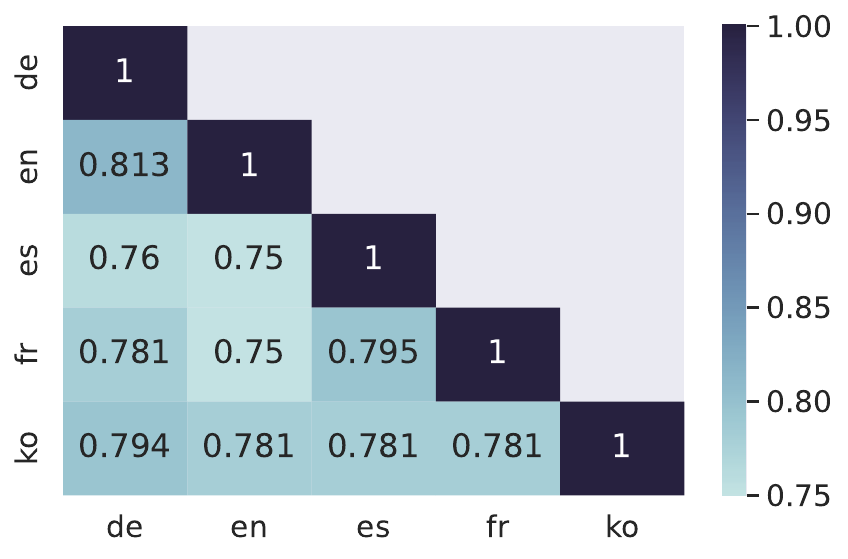}
     \includegraphics[width=0.49\linewidth]{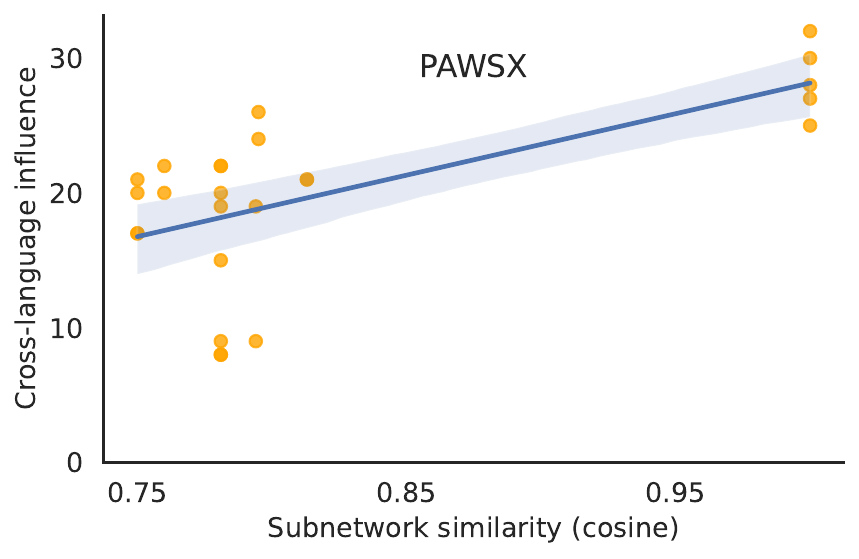}
      \includegraphics[width=0.49\linewidth]{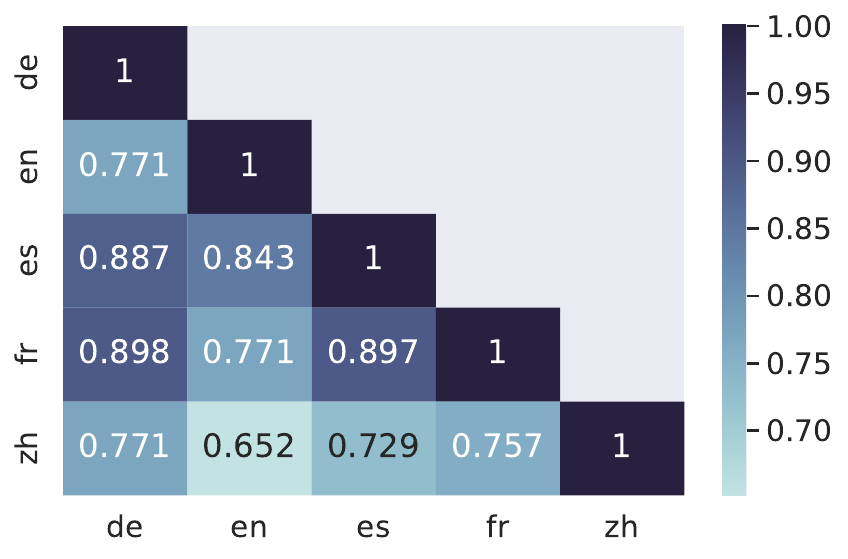}
     \includegraphics[width=0.49\linewidth]{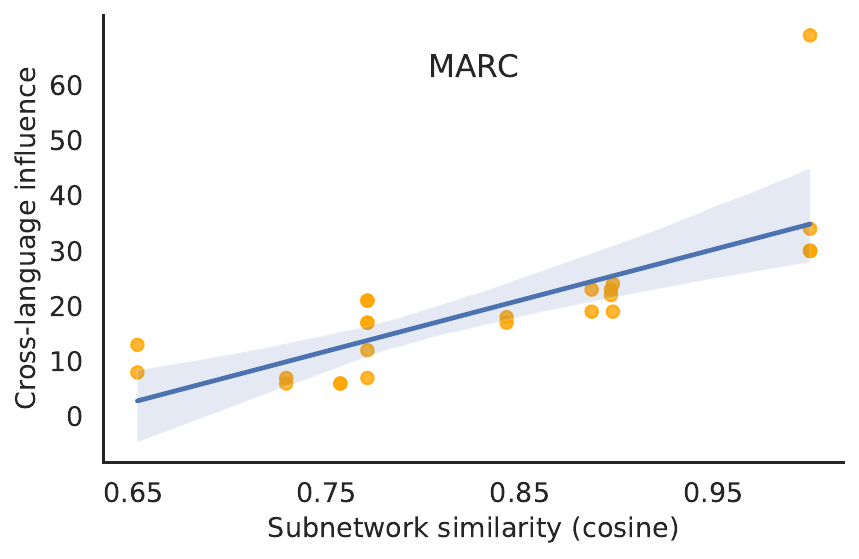}
    \caption{(\textbf{Left}) 
    The cosine similarity between the flattened binary subnetwork masks for each language pair.
    (\textbf{Right}) Positive cross-language influence as a function of structural (cosine) similarity between subnetworks.}
    \label{fig:subnetwork_corr}
\end{figure}
\begin{figure}[!t]
    \centering
   
     \includegraphics[width=0.5\linewidth]{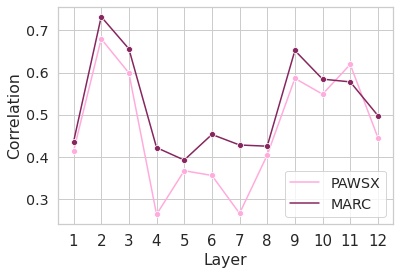}
    \caption{The correlation between positive cross-language influence and the subnetwork similarity computed based on individual model layers.}
    \label{fig:head_layer}
\end{figure}

\paragraph{Results} In Figure~\ref{fig:subnetwork_corr} (Left) we report the cosine similarity between the subnetworks of each language pair and (Right) the correlation between such subnetwork similarity and positive cross-language influence (in absolute numbers). From this, we find that for both tasks, subnetwork similarity is positively correlated with positive cross-language influence. 
Yet, we did not find a strong correlation between negative cross-language influence 
and subnetwork overlap. This is a promising finding, as it suggests that positive and negative influence do not necessarily have to go hand-in-hand. Thus, future work should investigate how we can further exploit subnetwork overlap to increase positive influence without increasing negative influence as well. 
Moreover, it is evident that for MARC the subnetworks show on average more overlap 
than for PAWS-X. Thus as the capacity within subnetworks from MARC have to be shared with more languages, it can explain why their language specialization is less strong as seen in Figure~\ref{fig:sft_diffs}. Future work should test whether SFT is still effective when using many more training languages (in which case subnetwork overlap will inevitably be higher).

\paragraph{Layer-wise analysis} To further analyze how subnetwork similarity affects cross-language influence, we now test how layer-wise subnetwork similarity correlates with performance. In Figure~\ref{fig:head_layer}, we see that similarity between certain layers is much more indicative of cross-language influence, and moreover, that both tasks follow very similar patterns despite ending up with vastly different subnetworks. This suggests that while language-specific subnetworks are also task-specific, there may be general language-specific properties across task-specific subnetworks that we can identify and exploit to better guide cross-lingual sharing.

\subsection{What happens within subnetworks during full model fine-tuning versus SFT?}
Previously, we used the sum of influence scores over model checkpoints to determine our rankings. To study how cross-language influence changes over time, given different training approaches, we now analyze the influence scores (and their rankings) from each checkpoint separately. 

 \begin{figure}[!t]
    \centering
    \includegraphics[width=\linewidth]{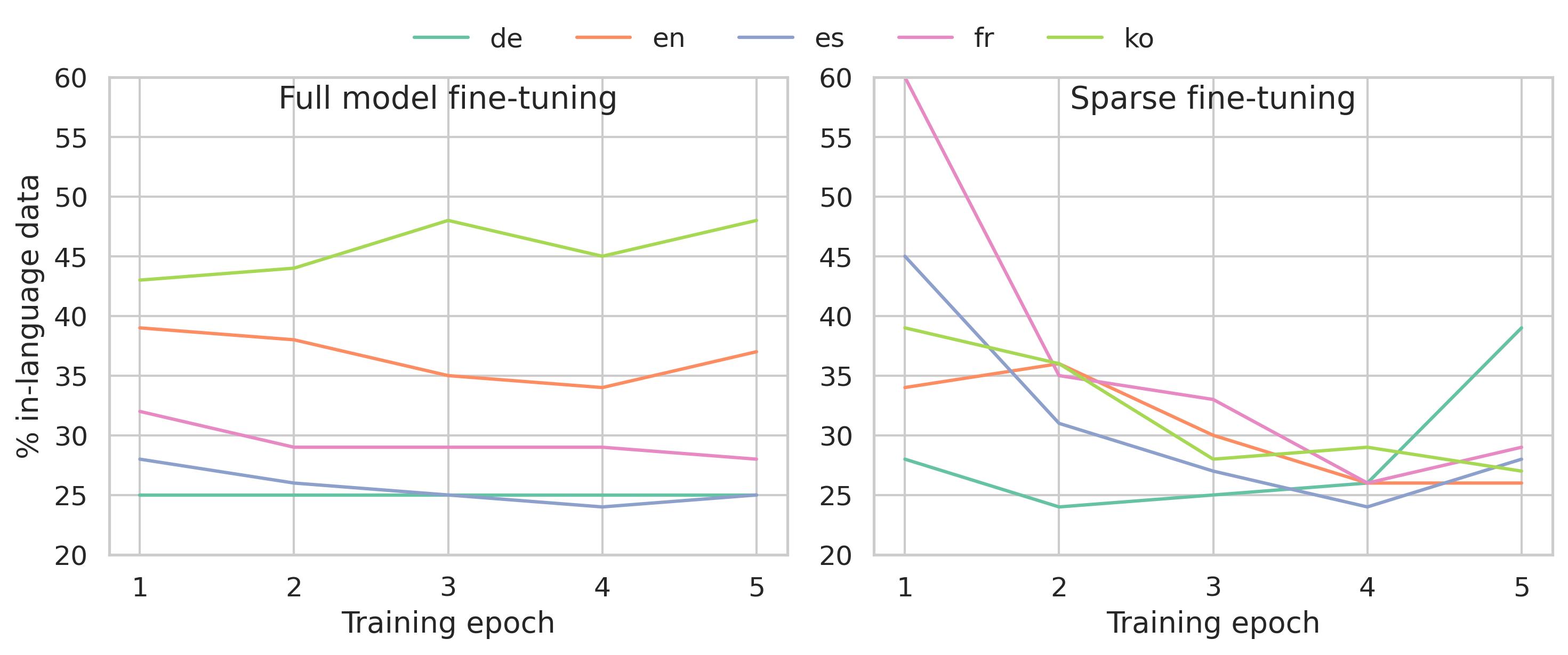}
    \caption{The change in language specialization of subnetworks over training epochs for PAWS-X.}
    \label{fig:epoch-change}
\end{figure}

\paragraph{Results} 
In Figure~\ref{fig:epoch-change}, we see that while both approaches converge to similar maximum levels of cross-lingual sharing ($\sim$25$\%$ reliance on in-language data) for PAWS-X, SFT allows \emph{all} training languages to share more data. Whereas for full model fine-tuning, Korean and English are left behind. 
Similarly, for MARC, Chinese is left behind, and using random subnetworks for PAWS-X has the same effect as full model fine-tuning (see Appendix~\ref{app:additional}). In line with results in previous sections
, we conclude that subnetworks must meaningfully overlap to improve cross-lingual interaction.

\section{Conclusion}
We studied to what degree modularity, in the form of language-specialized subnetworks, naturally arises within multilingual LMs. We demonstrate the existence of such subnetworks 
using 
TracIn to monitor the change in reliance on in-language data at test time when using subnetworks compared to the full model. Moreover, we studied the effects that SFT has on modularization, and find that it does not cause all subnetworks to become more specialized. Yet, in all cases, our identified subnetworks show vastly different behavior 
from random ones, indicating that we are able to uncover meaningful language-specific model behavior. 
Finally, 
we find that subnetwork similarity, particularly in specific model layers, correlates with positive, but not negative, cross-language influence. 
Future work should focus on further exploiting subnetworks and their interaction to better control cross-lingual sharing.

\section{Limitations}
One limitation of TDA methods in general is that the experiments are computationally expensive to run. While using the random projection method, explained in Appendix~\ref{app:efficiency}, somewhat mitigates the problem, it still prevents us from studying a wider range of LMs and/or larger models. Similarly, due to the computational costs, we are restricted to relatively easy tasks as (1) we can not use a large fine-tuning dataset and (2) TracIn operates on the sequence-level, i.e., it estimates how much a full training instance contributed to a prediction, making this method mostly suitable for classification and regression tasks. Given that the tasks are relatively simple, this might also limit the benefit of SFT over full model fine-tuning, hence the subnetwork behavior we see after SFT might be weaker than if we had studied more complex tasks. 

\section*{Acknowledgements}
We would like to thank Ian Tenney, Cindy Wang, and Tolga Bolukbasi for their feedback. This project was in part supported by a Google PhD Fellowship for the first author.



\clearpage
\appendix

\section{Low-memory sketches using random projections}\label{app:efficiency}
LMs have a large number of parameters which makes the inner product computations in the first-order approximation of the influence expensive, especially when computing influence scores for a large number of test points. Thus, following \citet{pruthi2020estimating}, we speed up the computations by using random projections, a method that allows us to pre-compute low-memory sketches of the loss gradients of the training points \citep{woodruff2014sketching} which can be stored and re-used to compute randomized unbiased estimators of the influence on different test points. To do so, we choose a random matrix $G \in \mathcal{R}^{d\times p}$, where $d \ll p$  is a user-defined dimension for the random projections, whose entries are sampled i.i.d.\ from $\mathcal{N}(0, \frac{1}{d})$ such that $E[G^TG] = \mathcal{I}$. 
Similarly, for the fully connected layers with a weight matrix $W\in \mathcal{R}^{m \times n}$, it is also possible to obtain a random projection of the gradient with respect to $W$ into $d$ dimensions. 
To do so, we use two independently chosen random projection matrices $G1 \in \mathcal{R}^{\sqrt{d}×m}$ and $G2 \in \mathcal{R}^{\sqrt{d}×n}$, where $E[G_1G^T_1 ] = E[G_2G^T_2] = I$, and compute:
\begin{equation}
    G_1 \nabla_y f(y)x^TG^T_2 \in \mathcal{R}^{\sqrt{d}×\sqrt{d}}
\end{equation}
, which can be flattened into a $d$-dimensional vector. See  Appendix E and F from \citet{pruthi2020estimating} for more details. Note that throughout our experiments we set $d=256$.

\section{Fine-tuning details}\label{app:fine-tuning}
For each task, we add a simple classifier on top of the pretrained XLM-R base model \citep{conneau2020unsupervised}. The classifier consists of one hidden layer and uses \texttt{tanh} activation. We then feed the hidden representation corresponding to the \texttt{<S>} token for each input sequence to the classifier for prediction. Moreover, following \citet{choenni2023languages}, we use AdamW \citep{loshchilovdecoupled} as an optimizer, and use learning rates of 2e-5, 9e-6, and 2e-5 for XNLI, PAWS-X and MARC respectively. 

\section{Baseline results}
\label{sec:full_baseline}

\begin{figure}[H]
    \centering
      \subcaptionbox{XNLI}{
       \includegraphics[width=\linewidth]{ 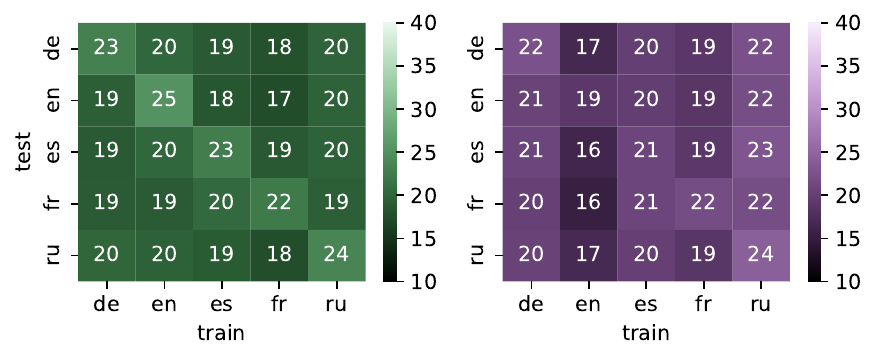}
       \vspace{-0.2cm}} 
     \subcaptionbox{PAWS-X}{
    \includegraphics[width=\linewidth]{ 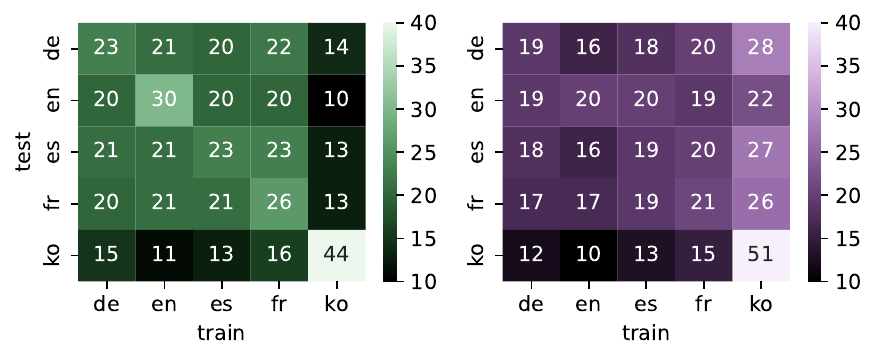}
        \vspace{-0.2cm}}      
        \subcaptionbox{MARC}{
   \includegraphics[width=\linewidth]{ 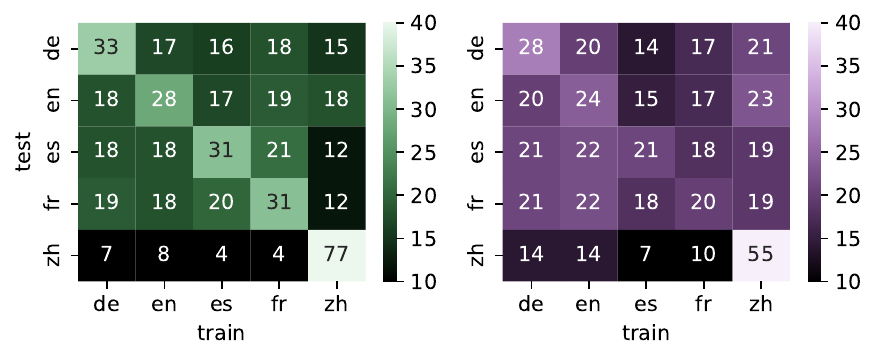}
       \vspace{-0.2cm}}  
\caption{Percentage that each training language contributes to the top 100 training samples for each test language when using the full model. Results are averaged over all 500 test samples per language.}
    \label{fig:knn_distr}
\end{figure}

\section{Details on the identified subnetworks}

\begin{figure}[H]

    \centering
    \subcaptionbox{\footnotesize{XNLI}}{
      \includegraphics[width=0.85\linewidth]{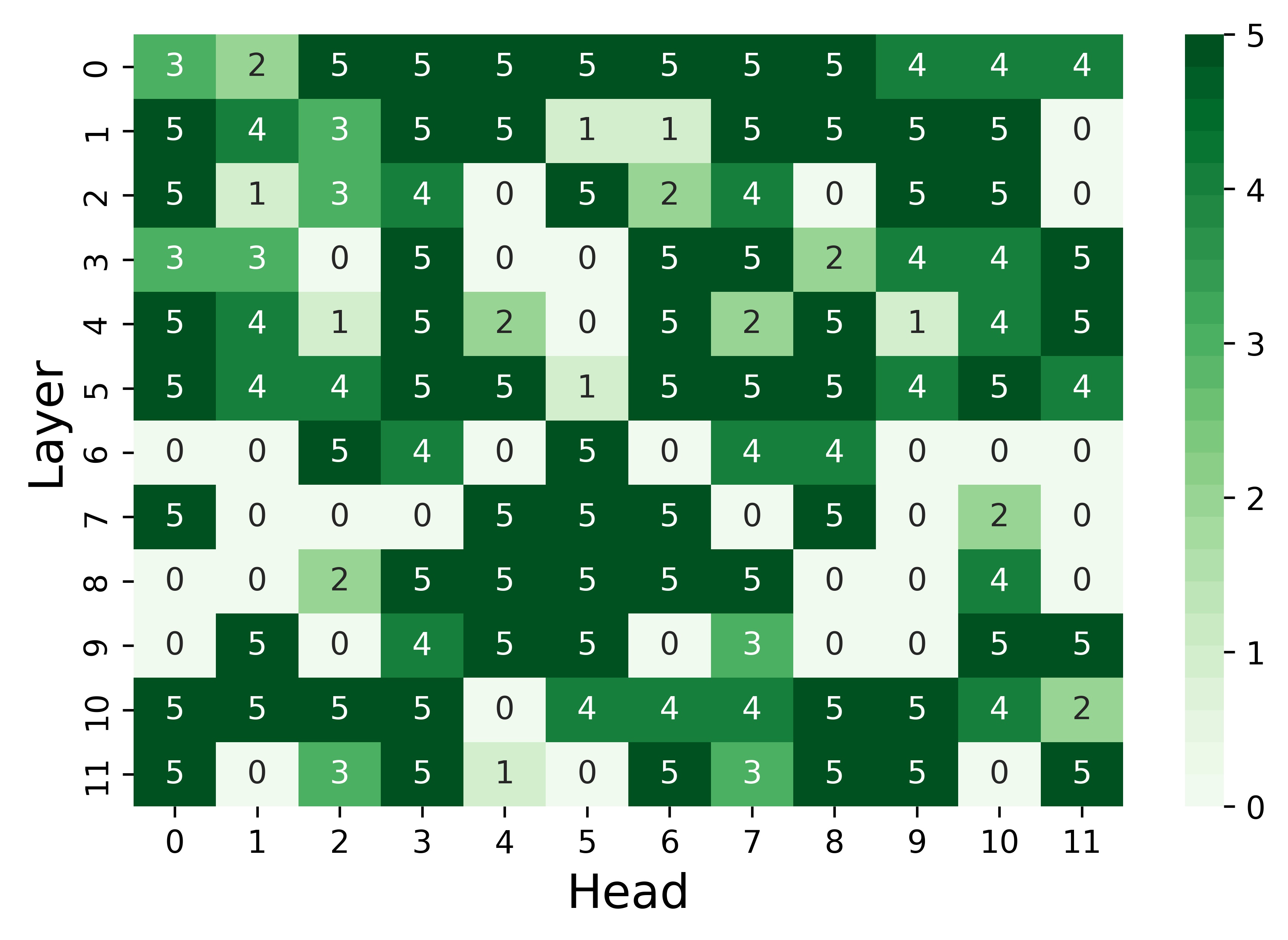}}
         \subcaptionbox{PAWS-X}{
     \includegraphics[width=0.85\linewidth]{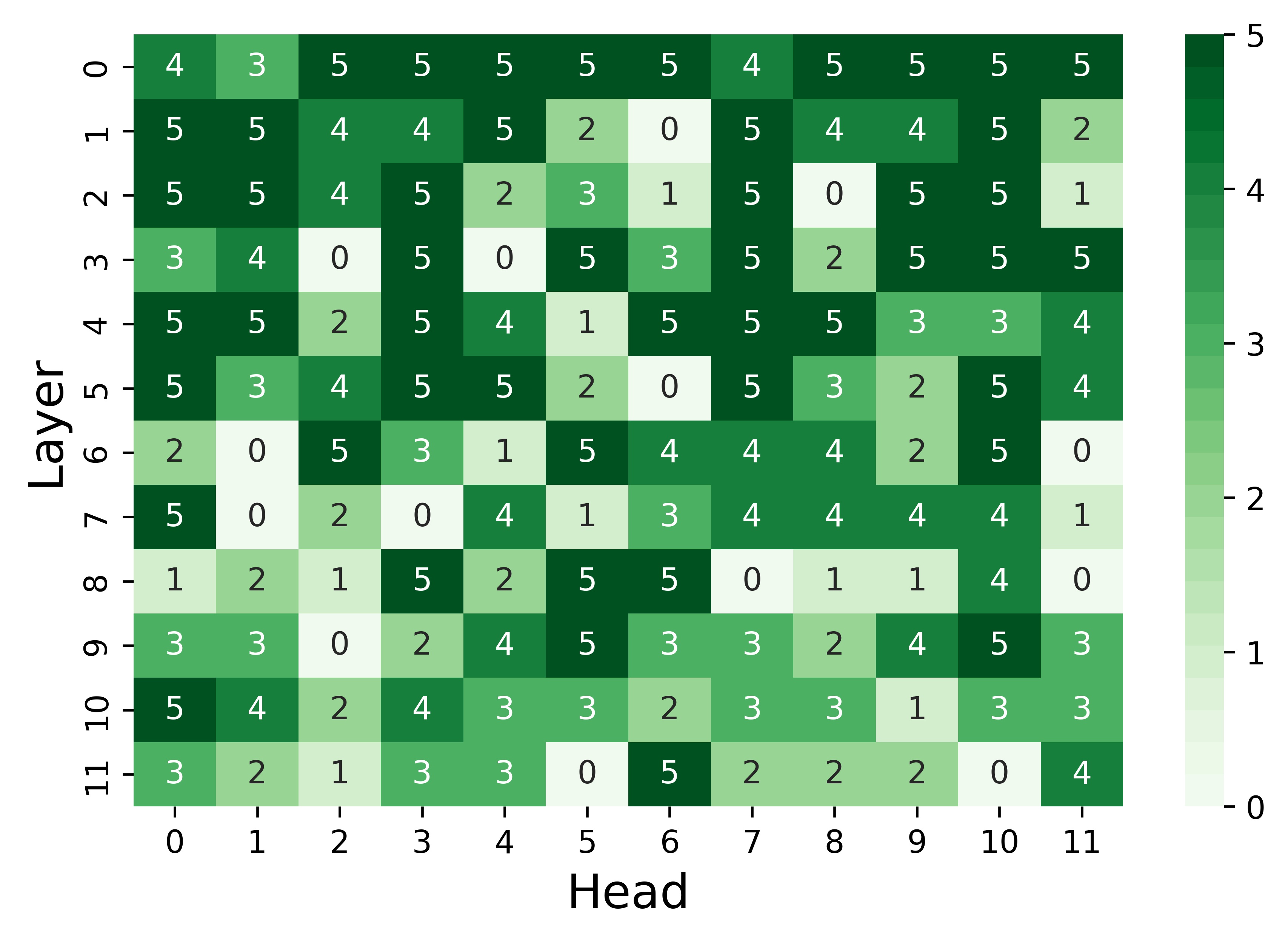}}
        \subcaptionbox{MARC}{
          \includegraphics[width=0.85\linewidth]{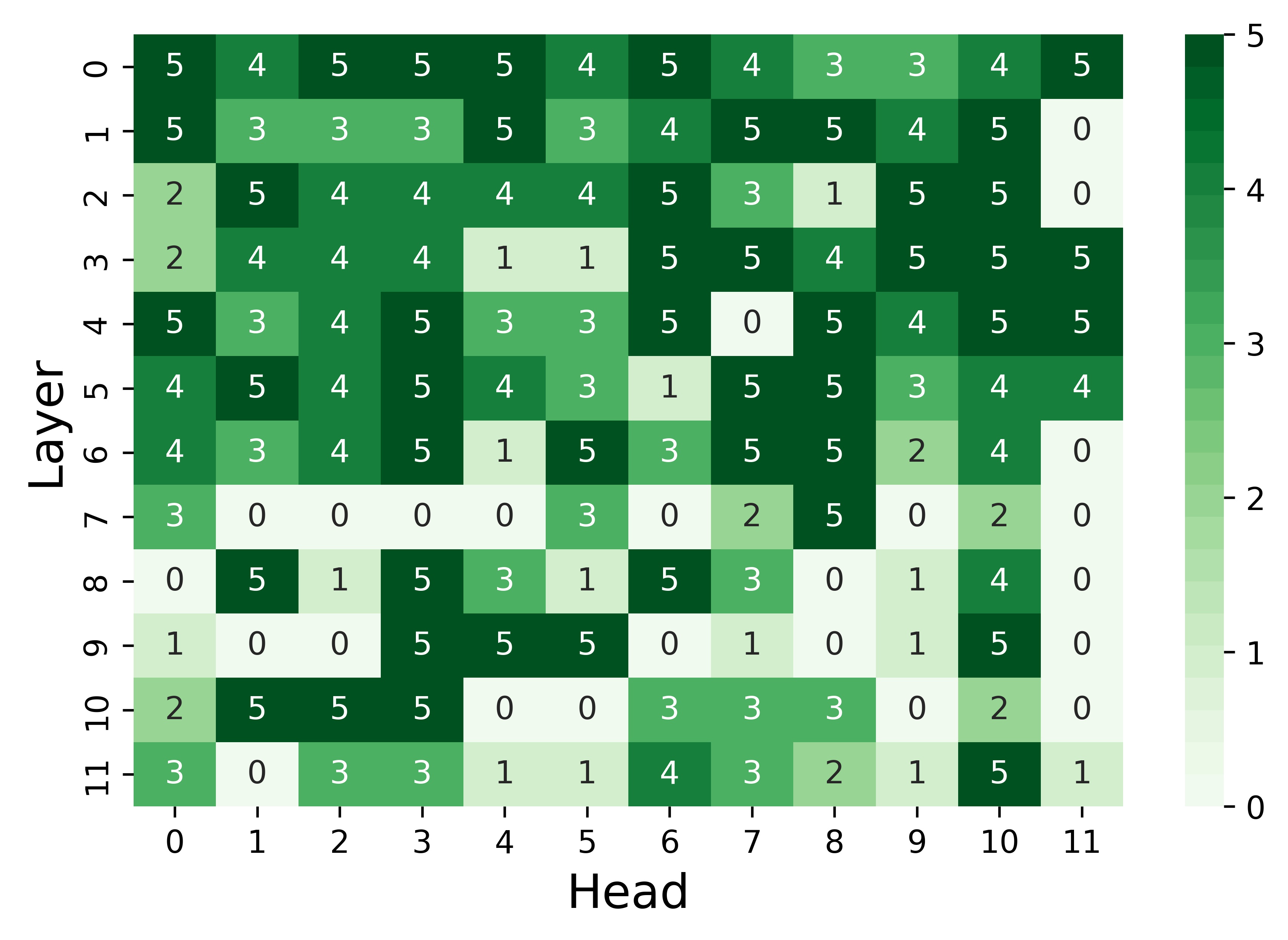}}
     \caption{The overlap of heads enabled by each language's subnetwork per task. 5 indicates that the head is shared across all languages and 0 that it is not used by any of the languages.}
         \label{fig:overlap}
  \end{figure}

\begin{figure}[H]

    \centering
      \includegraphics[width=0.49\linewidth]{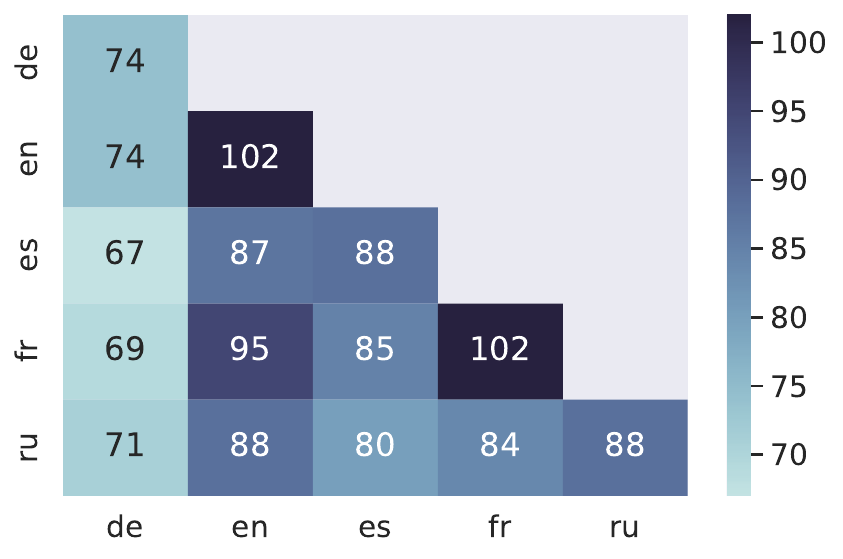}
          \includegraphics[width=0.49\linewidth]{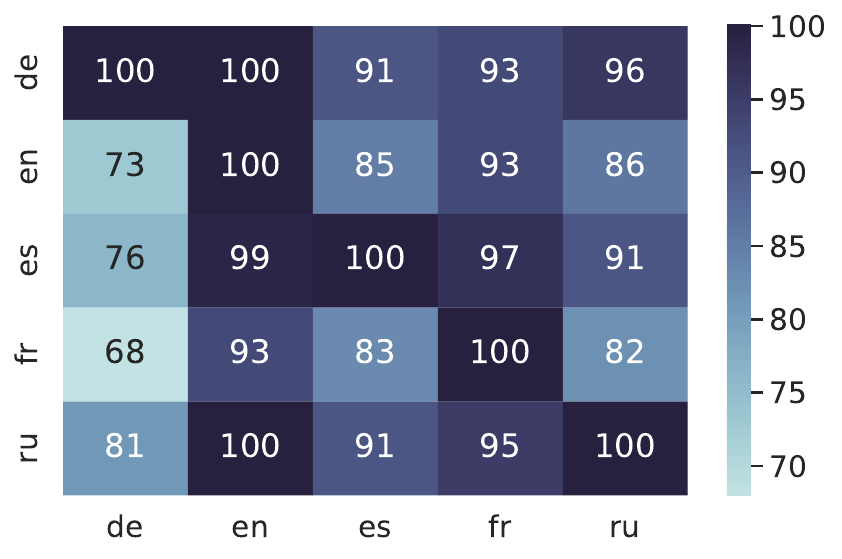}
     \caption{The absolute number of overlapping attention heads between each language pairs' subnetworks for XNLI. \textbf{(Left)} The percentage of overlap in heads between each language pairs' subnetworks. Note that values are not symmetric between language pairs as each language's subnetwork can have a different sparsity level. For instance, for German on the $y$-axis, it shows that 100$\%$ of the enabled heads are shared with English. Yet, 73$\%$ of the enabled heads for English are shared with German, given that English has more heads enabled. 
     \textbf{(Right)}}
         \label{fig:overlap_xnli_count}
  \end{figure}

\begin{figure}[H]

    \centering
      \includegraphics[width=0.49\linewidth]{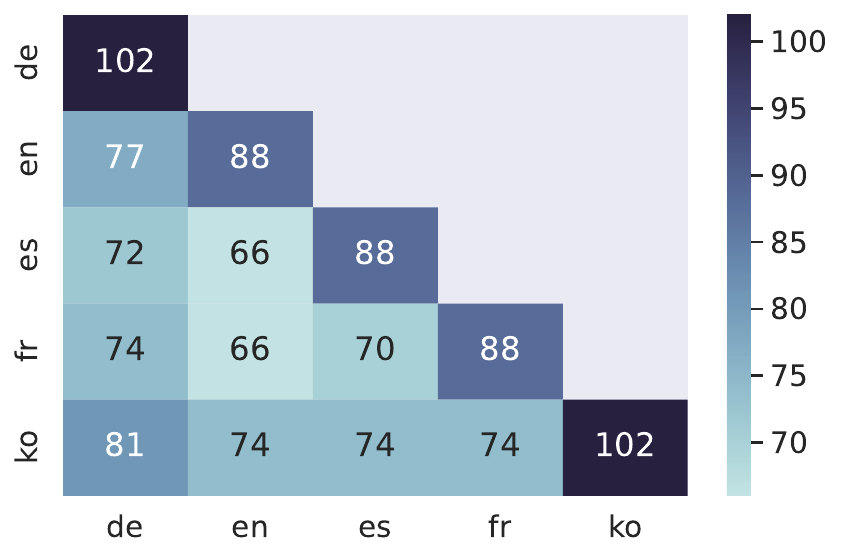}
          \includegraphics[width=0.49\linewidth]{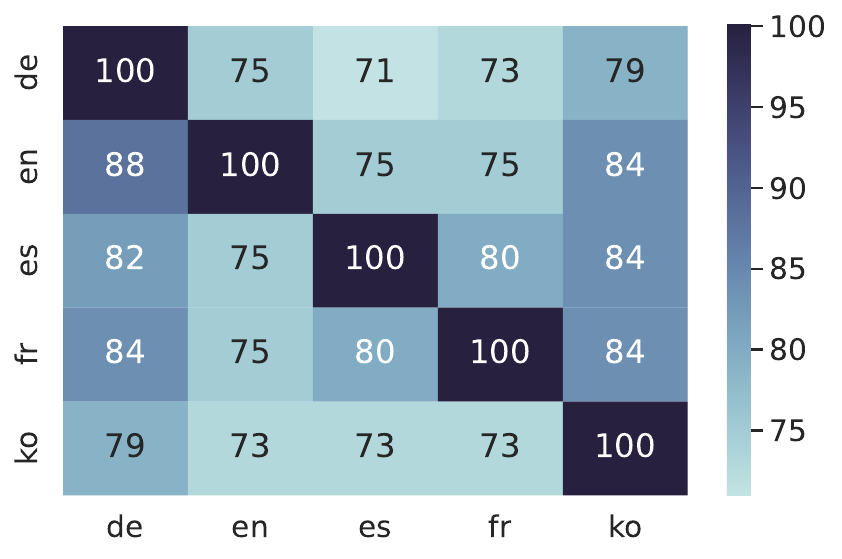}
        
     \caption{The absolute number of overlapping attention heads between each language pairs' subnetworks for PAWS-X. \textbf{(Left)} The percentage of overlap in heads between each language pairs' subnetworks. Note that values are not symmetric between language pairs as each language's subnetwork can have a different sparsity level. For instance, for German on the $y$-axis, it shows that 75$\%$ of the enabled heads are shared with English. Yet, 88$\%$ of the enabled heads for English are shared with German, given that English has fewer heads enabled. 
     \textbf{(Right)}}
         \label{fig:overlap_count}
  \end{figure}

\begin{figure}[H]

    \centering
      \includegraphics[width=0.49\linewidth]{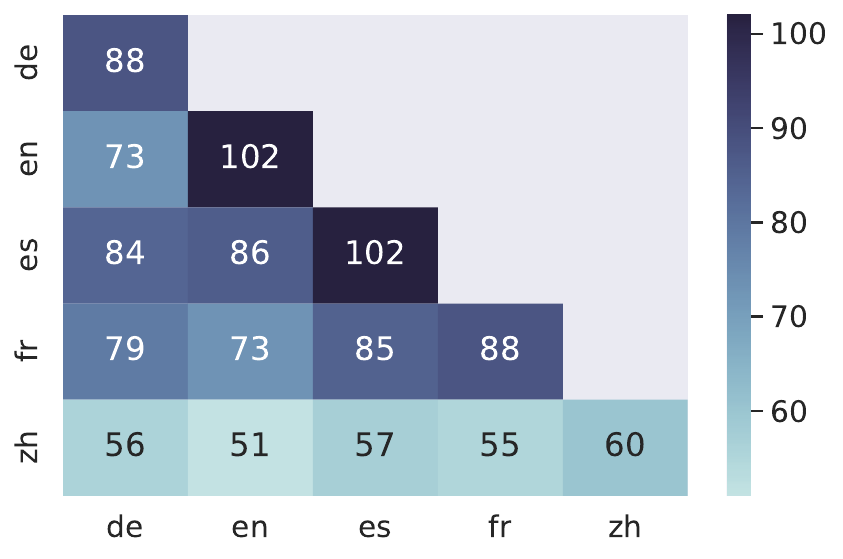}
        \includegraphics[width=0.49\linewidth]{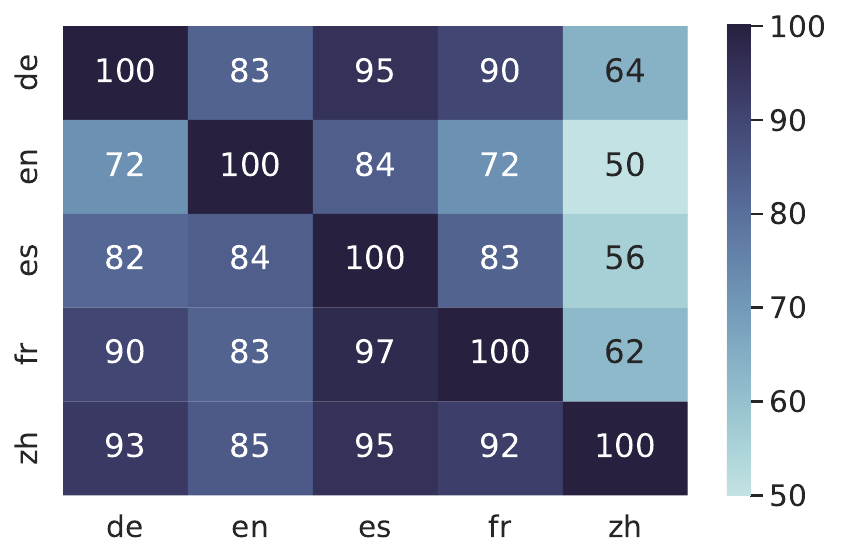}
     \caption{The absolute number of overlapping enabled heads between each language pairs' subnetworks for MARC. \textbf{(Left)}  The percentage of overlap in heads between each language pairs' subnetworks. Note that values are not symmetric between language pairs as each language's subnetwork can have a different sparsity level. 
     \textbf{(Right)}}
         \label{fig:overlap_marc_count}
  \end{figure}

In Figure~\ref{fig:overlap}, we show the overlap in attention heads of the identified subnetworks that we found for each of our 5 training languages. While we find that all subnetworks have similar sparsity levels (see Table~\ref{tab:sparsity} for the absolute number of disabled attention heads per task and language), we also see that across all tasks, some heads are not used by any of the languages (indicated by 0).  
This finding suggests that the model capacity does not have to be a limiting factor within this model, as more language-specific parameters could be assigned if needed. In contrast, many heads, especially in the lower layers of the models for PAWS-X and in the higher layers for XNLI, are fully shared across all languages. Given that paraphrasing relies more on lower-level syntactic information than NLI, this is in line with previous findings that suggest that syntax is processed in lower layers while semantics in processed in the higher ones \citep{tenney2019bert}. Moreover, in Figures~\ref{fig:overlap_xnli_count},~\ref{fig:overlap_count} and~\ref{fig:overlap_marc_count}, we see for XNLI, PAWSX-X and MARC the amount of subnetwork overlap between each language pair both in absolute values and as a percentage of the language's full subnetwork capacity.

\label{app:subnet_details}
    \begin{table}[H]
    \centering
    \begin{tabular}{c|ccccccc}
             &  de & en & es & fr & ko & ru & zh\\
             \hline
         PAWS-X& 42& 56&56 &56 &42 & -&- \\
        XNLI&  70&42 &56 & 42& -& 56&- \\
         MARC & 56&42 & 42& 56&- &- &84 \\
    \end{tabular}
    \caption{The number of disabled attention heads in the identified subnetwork of each language and task.}
    \label{tab:sparsity}
\end{table}


\section{Additional results}\label{app:additional}
\begin{figure}[H]
    \centering
     \includegraphics[width=0.49\linewidth]{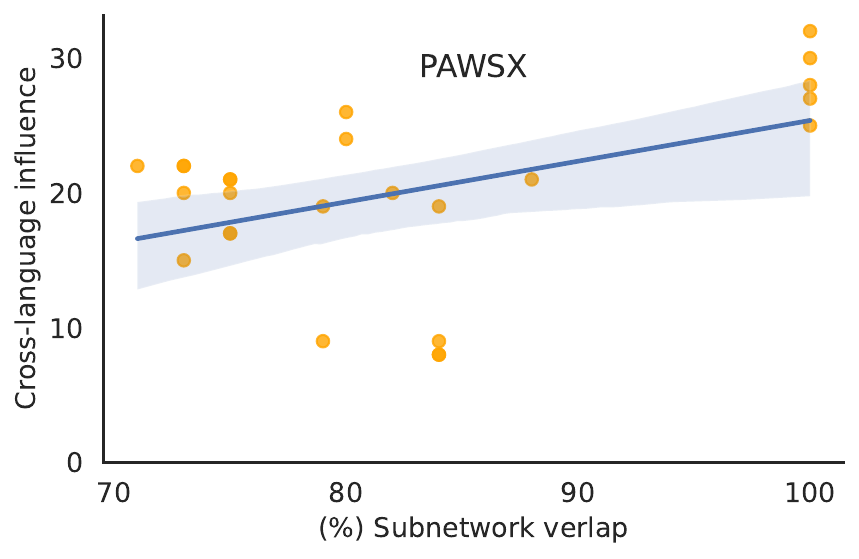}
       \includegraphics[width=0.49\linewidth]{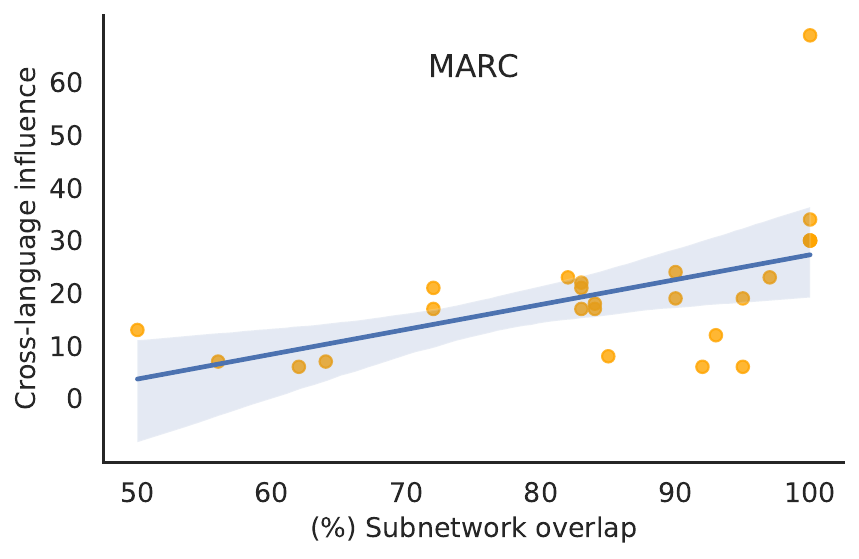}
    \caption{The correlation between the percentage of overlap in heads between each language pairs' subnetworks and their amounts of cross-language interference (in absolute numbers).}
    \label{fig:enter-label}
\end{figure}

\begin{table}[H]
    \centering
    \begin{tabular}{c|cc|cc}
    & \multicolumn{2}{c|}{PAWS-X} & \multicolumn{2}{c}{MARC}\\
    &  Full & SFT & Full & SFT  \\
    \hline
        de &  68.0 & 78.8 & 75.3 & 76.4 \\
        en &  78.6& 83.0 & 75.1 & 75.8 \\
        es & 78.2& 80.5 & 76.6 & 77.4\\
        fr &  82.1& 79.8 & 76.2 & 77.6\\
        ko & 67.1& 69.9 & -- & --\\
        zh & -- & -- & 69.5 & 71.1\\

    \end{tabular}
    \caption{The performance effect of SFT compared to full model fine-tuning. We report the performance of the language-specific subnetworks when used on the test samples from the respective languages when using either one of the fine-tuning techniques. Note that we do not optimize for obtaining SOTA performance in this study e.g.,\ we train on relatively little data to make our TracIN experiments computationally feasible.}
    \label{tab:performance}
\end{table}

\begin{figure}[H]
  \includegraphics[width=\linewidth]{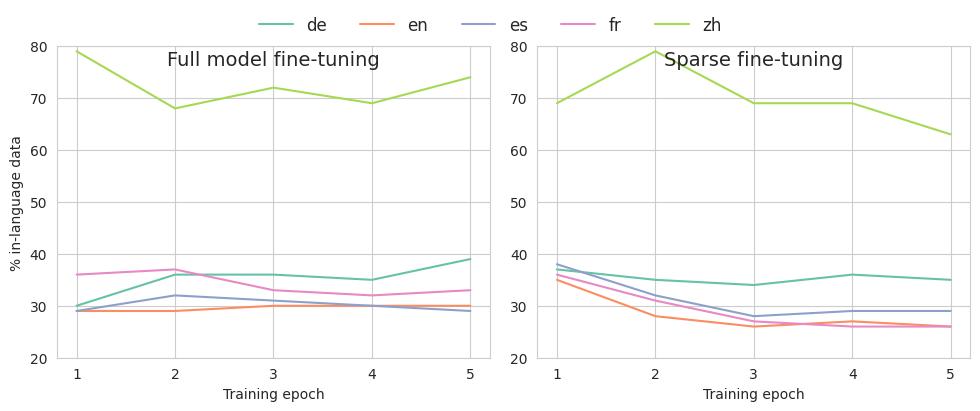}
    \caption{The change in language specialization for each test language over training epochs for MARC. We see that the patterns for full model fine-tuning are similar to PAWS-X, yet for sparse fine-tuning they differ considerably.}
    \label{fig:epoch-marc}
\end{figure}

\begin{figure}[H]
    \centering
    \includegraphics[width=0.5\linewidth]{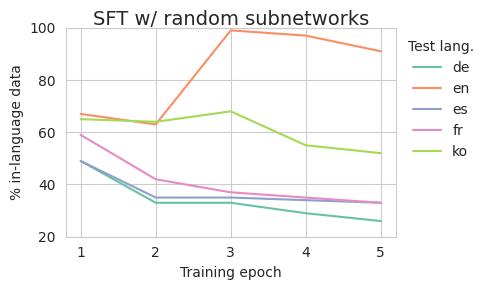}
    \caption{The language specialization effect of SFT with random subnetworks on PAWS-X over training epochs. }
    \label{fig:sftwrandom-epochs}
\end{figure}

\section{Additional experiments}

\subsection{Composing subnetworks at test time}
As an additional analysis, we study whether we can compose two languages' identified subnetworks into a language-pair specific subnetwork that, when applied at test time, will enforce more cross-lingual reliance on each other's training data. For merging two subnetworks we both tried taking the union and the intersect of the respective binary subnetwork masks. Note that we apply the composed subnetwork only at test time to a model that was trained with SFT (using the initial identified subnetworks).

\paragraph{Results}

\begin{figure}[!t]
    \centering
    \includegraphics[width=0.49\linewidth]{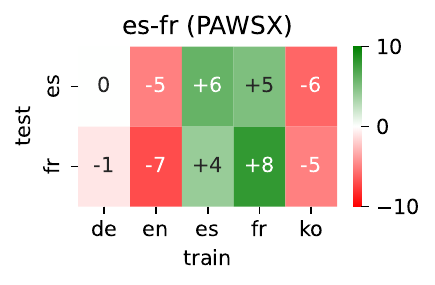}
    \includegraphics[width=0.49\linewidth]{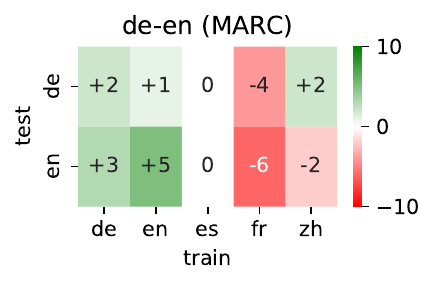}
    \caption{The effect on the contribution of positive influence from each training language when composing two language's subnetworks by their intersect and applying them at test time (compared to full model fine-tuning).}
    \label{fig:intersect}
\end{figure}

We find that we can only successfully enforce cross-lingual sharing through subnetwork composition for two languages, if those individual language's subnetworks already stimulated cross-lingual sharing between the pair. For instance, in Figure~\ref{fig:sft_diffs}, we saw that both the Spanish and French subnetworks (PAWS-X) and the German and English ones (MARC) resulted in more sharing between the pairs. In Figure~\ref{fig:intersect}, we show that taking the intersections of those language pairs' subnetworks can further strengthen this behavior (taking their union resulted in sharing to a lesser extent) 
Trying to control sharing behavior by composing two language-specific subnetworks that individually did not lead to more sharing between the pair did not yield any clear positive results. This shows that while SFT can better cross-lingual sharing, there is still much room for improvement when it comes to creating a truly modular system that enables compositionality. 

\end{document}